\pdfoutput=1

\documentclass[11pt]{article}

\usepackage{acl}

\usepackage{times}
\usepackage{latexsym}
\usepackage{url}
\usepackage{hyperref}
\usepackage{graphicx}
\usepackage{booktabs}
\usepackage{paralist, multirow}
\usepackage{cleveref}

\usepackage[T1]{fontenc}

\usepackage[utf8]{inputenc}

\usepackage{microtype}

\usepackage{inconsolata}

\title{Connecting the Dots in News Analysis: Bridging the Cross-Disciplinary 
Disparities in Media Bias and Framing}

\author{Gisela Vallejo$^1$ \qquad Timothy
  Baldwin$^{1,2}$ \qquad Lea Frermann$^1$ \\
  $^1$The University of Melbourne \qquad $^2$MBZUAI \\
  \texttt{gvallejo@student.unimelb.edu.au},\\
 \texttt{{\{tbaldwin,lfrermann\}@unimelb.edu.au}}
}

\begin{document}
\maketitle
\begin{abstract}
The manifestation and effect of bias in news reporting have been central topics in the social sciences for decades, and have received increasing attention in the NLP community recently. 
While NLP can help to scale up analyses or contribute automatic procedures to investigate the impact of biased news in society, we argue that methodologies that are currently dominant fall short of capturing the complex questions and effects addressed in theoretical media studies. This is problematic because it diminishes the validity and safety of the resulting tools and applications. 
Here, we review and critically compare task formulations, methods and evaluation schemes in the social sciences and NLP.
We discuss open questions and suggest possible directions to close identified gaps between theory and predictive models, and their evaluation. These include model transparency, considering document-external information, and cross-document reasoning.
\end{abstract}

\section{Introduction}

The depiction of complex issues in the media strongly impacts public opinion, politics, and policies~\citep{ghanem-1997-filling,giles-2009-psychology}.
Because a handful of global corporations own an increasing proportion of news outlets, the reach and impact of biased reporting are amplified~\cite{hamborg-2020-media}. 
Although perfect neutrality is neither realistic nor desirable, media bias turns into an issue when it becomes systematic. 
If the public is unaware of the presence of bias, this can lead to dangerous consequences, including intolerance and ideological segregation \citep{baly-etal-2020-detect}.

\begin{figure}
    \centering
    \includegraphics[scale=0.33]{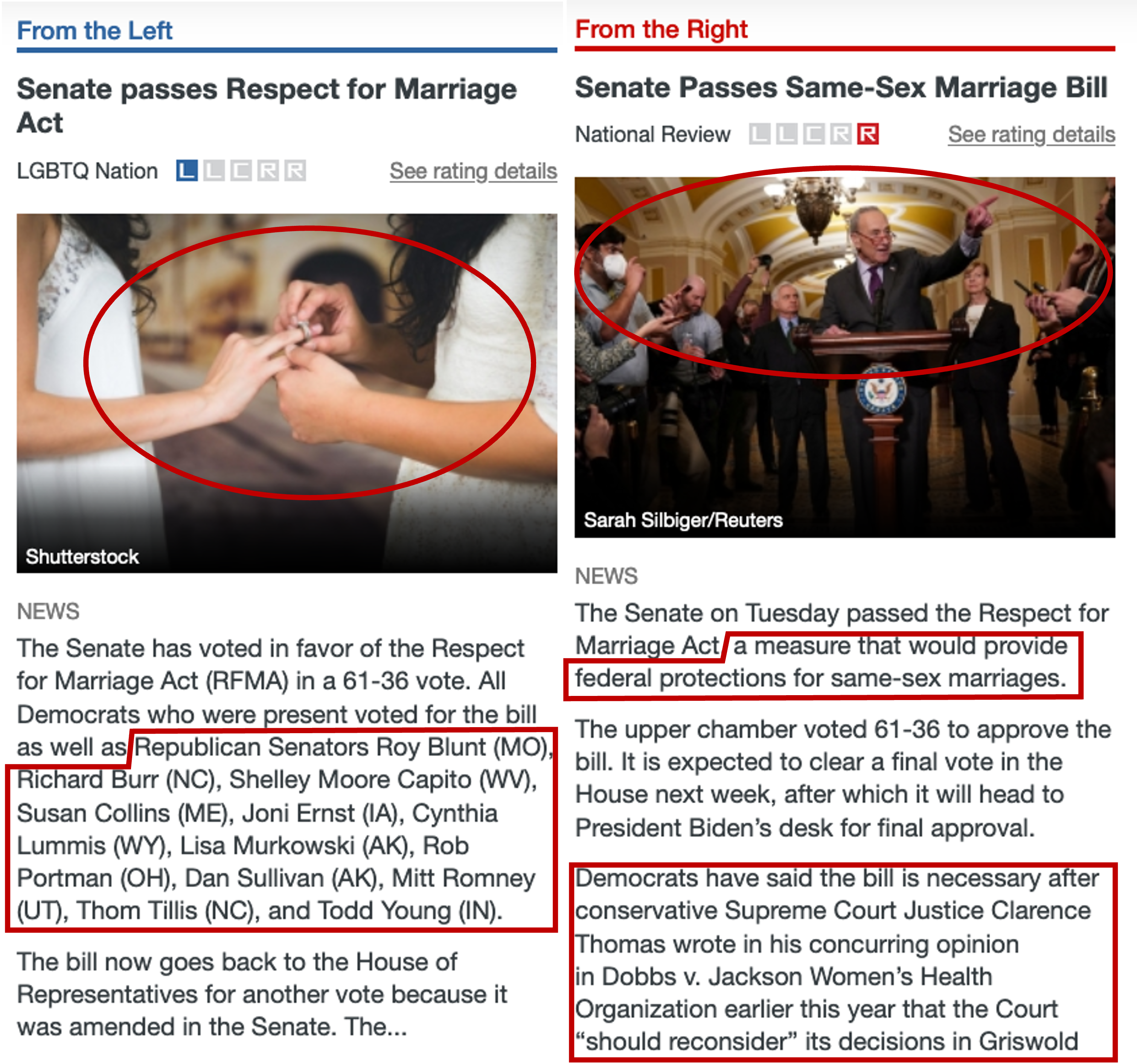}
    \caption{Two articles about the same event written from different political ideologies (Source: \url{allsides.com}).}
    \label{fig:allsidesimg}
\end{figure}

Figure~\ref{fig:allsidesimg} illustrates the concepts of {framing} and {media bias} adopted in this paper, using the passing of the Respect for Marriage Act as an example. 
{\it Framing} refers to the emphasis of selected facts with the goal of eliciting a desired interpretation or reaction in the reader~\citep{entman-2007-framing}. 
The left-leaning article in Figure~\ref{fig:allsidesimg} leads with an uplifiting picture of a wedding and emphasizes bill support, evoking a positive framing of new opportunities for same-sex couples; while the right-leaning article focuses on disputes in both image and text, framing the issue in a more negative light. 
\textit{Political bias} refers to partisan slanted news stories, or the ``tendency to deviate from an accurate, neutral, balanced, and impartial representation of `reality' of events and social world''~\citep{mcquail-2020-mcquails}, which can be a result of a selected framing.
In \Cref{fig:allsidesimg}, each document was flagged as far-left and far-right ideological leaning, respectively, on the basis of their publishing media outlets.
Political bias is typically deliberate~\citep{williams-1975-unbiased} while framing may be inadvertent and caused by external pressures such as space limitations.


Framing and media bias have been under active research in different subfields of the social sciences. Angles of study include the manifestation of frames in the mass media and their effects on public opinion (communication sciences); the impact of frames in groups' and individuals' sensemaking of the world (social psychology; sociology) or on their observable  behaviour (economics and political science). We focus on the first notion: systematic analyses of framing bias in the mass media, through manual coding, or with NLP technology. 
In this paper, we will collectively refer to the studies of communication and mass media as social sciences.

With the increasing pace and almost complete digitisation of news reporting there is a need and opportunity to scale the analysis of media bias~\cite{parasie-2022-computing}. Besides, evidence suggests that exposing media bias promotes healthy public debate, aids journalists to increase thoroughness and objectivity, and promotes critical news consumption~\citep{dallmann-etal-2015-media}. We discuss the specific role of NLP in this context in \Cref{sec:discussion}.

\subsection{Contribution and Approach} %
\label{ssec:approach}
We relate the NLP research landscape on framing and media bias prediction to typical research questions and hypotheses in the social sciences. We tease out disconnects across disciplines, and make concrete suggestions on how social science approaches can improve NLP methodology, and how NLP methods can more effectively aid social science scholars in their analyses and underpin technology to raise awareness of media bias.

\citet{hamborg-etal-2019-automated} present an overview of traditional and computational approaches to media bias, including detailed definitions of bias types and their emergence in the context of news production. 
We complement this survey by contextualising recent approaches in NLP with dominant questions and approaches in the social sciences. 
\citet{ali-hassan-2022-survey} review computational approaches to modelling framing providing a systematic overview of NLP and machine learning methods. 
In contrast, we critically review the methodological decisions along the higher-level NLP pipeline: data (\Cref{sec:data}), problem formulation (\Cref{sec:method}), and evaluation (\Cref{sec:metrics}), link them back to social science methodology, and pinpoint gaps between the two disciplines. We motivate our focus with a case study in \Cref{sec:casestudy}.

We obtained an up-to-date inventory of NLP approaches to media bias and framing, as well as a representative body of corresponding work in the relevant social science disciplines as follows. We collected relevant NLP benchmark data sets (Table~\ref{tab:datasets}) and the papers that addressed them for a broad-coverage overview of approaches in the field. We complement this with social science papers departing from citations in \citet{hamborg-etal-2019-automated}. Here, we do {not} attempt a systematic literature review, but rather present a {\it representative} body of work across the fields.\footnote{We intentionally depart from the traditional approach of selecting the top $N$ results from a research anthology for a few simple queries, as this would would not capture the diversity of works both in terminology and publication venues.} 
We excluded papers that a) duplicated methodologies, b) provided redundant definitions, or c) focused on unrelated topics. From this selection process, our final corpus comprises 63 papers (36 framing, 27 media bias), which were considered for further analysis, also listed in~\Cref{sec:appendix}.

\section{Background: Framing and Media Bias}
{\it Framing} and {\it politically biased news reporting} are two strategies to systematically promote specific perspectives on contested issues. 
We note that not every presence of framing is political bias and not all political bias is represented as framing but their intersection can reinforce each other’s impact. 
They are overlapping concepts which have been addressed jointly or with similar methods in NLP. As such, we include both strategies in this survey.

{\it Framing} has been conceptualised variously in different social science disciplines. Prevalent notions of framing include \textit{equivalence framing} -- presenting the same logical information in different forms \citep{cacciatore2016end} -- and \textit{emphasis framing} -- highlighting particular aspects of an issue to promote a particular interpretation \citep{entman-2007-framing}. 
Additionally, framing has been conceptualised as a process \citep{devreese-2005-news,entman-2007-framing,chong-druckman-2007-framing}, a communication tool \citep{scheufele-1999-framing}, or a political strategy \citep{roy-goldwasser-2020-weakly}. 
Frames have been conceptualised within different dichotomies. \citet{devreese-2005-news} distinguishes \textit{issue-specific} and {\it issue-generic} frames which apply to only a single or across several issues, respectively. \citet{scheufele-1999-framing} differentiates between \textit{media frames}, as embedded in the political discourse, and \textit{audience frames}, as the reader's interpretation of an issue. 
Finally, \citet{iyengar_1991_anyone} defines \textit{episodic framing} as portraying an issue with an individual example compared to \textit{thematic framing}, which takes broader context into account. 
Here, we cover both issue-specific and issue-generic frames and attach to \citet{entman-2007-framing}'s notion of emphasis framing.

While {framing} is a priori detached from partisan views, {\it political bias} refers to an explicit association of an article or media outlet with a specific political leaning. Both concepts result in biased news reporting, and correspondingly
NLP researchers have attempted to address them jointly, either by investigating political framing \citep{roy-goldwasser-2020-weakly} or by identifying correlations between framing and partisan slanted articles \citep{ziems-yang-2021-protect-serve}. 
NLP studies have attempted automatic media bias identification under several names, including: hyper-partisan news detection \citep{kiesel-etal-2019-semeval}, media bias detection \citep{spinde-etal-2021-neural-media,lei-etal-2022-sentence}, identification of biased terms \citep{spinde-etal-2021-identification}, and political ideology detection \citep{iyyer-etal-2014-political,kulkarni-etal-2018-multi}. Their common goal is to detect and classify the bias of a data sample towards a particular political ideology. 
Many of these approaches naturally relate to investigating \emph{how the story is told} (i.e., framing).  

\section{Three Disconnects}
\label{sec:casestudy}
To illustrate the disconnects between the social sciences and NLP, we use a representative study of media bias from the communication sciences~\cite{hernandez-2018-killed} which investigates the framing of domestic violence in the South China Morning Post. The author formulates two research questions:
\begin{compactenum}
\item Framing functions: Are femicides recognized as a problem of domestic violence? What are their causes, and the solutions proposed?
\item Frame narratives: What are the main narratives? And which sources are cited in support?
\end{compactenum}

\begin{figure}
    \centering
    \includegraphics[scale=0.33]{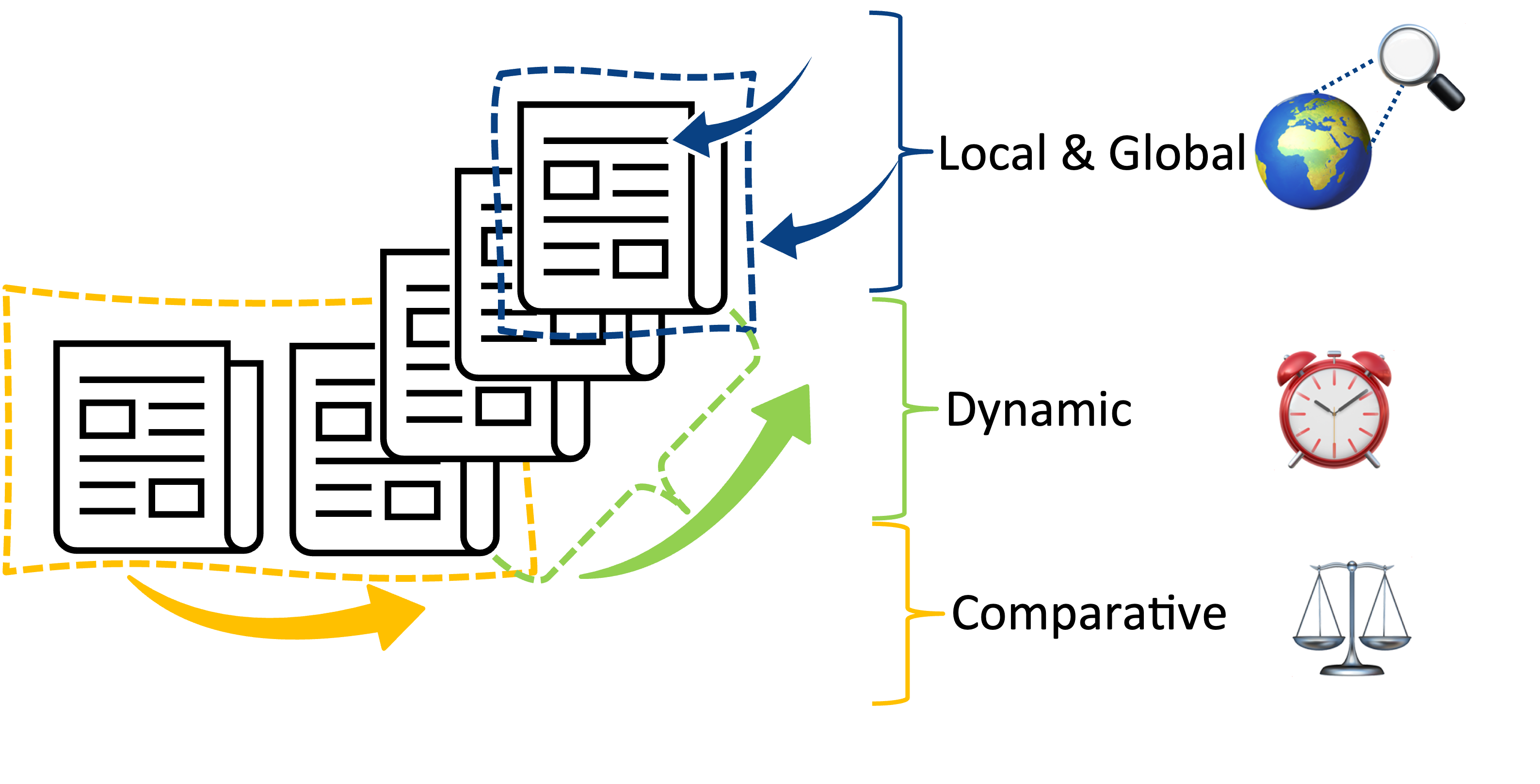}
    \caption{Illustration of the three disconnects: framing is both local and global (blue), dynamic (green) and best identified through comparative analysis (yellow).}
    \label{fig:disconnects}
\end{figure}

The first research question considers the {\it local} written aspects within each news article. Specifically, it studies the causes and solutions presented, grounded in \citet{entman-1993-framing}'s conceptualisation of framing in terms of a problem, its cause, and its solution. The second research question relates these local aspects to a {\it global} (cross-document) view by contrasting narratives that present domestic violence as isolated incidents with those that treat it as a societal problem. It further connects the articles to {\it extrinsic} variables, including the sources used and cultural contexts of the story (e.g.\ whether the article refers the role of women in the Chinese family or understands domestic violence through the lens of the Confucian philosophy). Furthermore, the study considers articles over an extended period, capturing the {\it temporal development} of framing and bias. In contrast, current NLP approaches to frame prediction have predominantly adopted a single-label prediction approach per unit of analysis~\cite{baumer-etal-2015-testing,naderi-hirst-2017-classifying,liu-etal-2019-detecting}, rather than treating frames as  structures which could decompose into aspects like cause vs.\ solution (but see~\citet{akyurek-etal-2020-multi, mendelsohn-etal-2021-modeling,frermann-etal-2023-conflicts} for recent exceptions). Current approaches furthermore treat units of analysis (sentences, articles) as independent without considering links across documents, across time, or to document-external context. 
The multi-level and dynamic understanding of bias and framing is fundamental in the social science studies. 
In sum, we identify three fundamental properties of bias and framing that underpin social science research on bias and framing, and we also visually represent them in~\Cref{fig:disconnects}:

\paragraph{Framing/bias is local and global} It is local, because an article can contain several frames, and it is global because understanding the framing of an article may require to aggregate local frames and link them with information such as cited (or omitted) sources, or the outlets' political leaning.

\paragraph{Framing/bias is dynamic} Frames change across time, outlets, countries, and communities. Understanding the {\it dynamics} of framing can shed light on trends and the impact of a sustained exposure to biased reporting on readers' opinions. 
 
\paragraph{Framing/bias as a comparative task} Media bias and framing are most apparent when directly contrasting articles from different perspectives, places or times (cf., Figure~\ref{fig:allsidesimg}). Formulating our task in a comparative way -- rather than predicting instance labels in isolation -- may improve the quality, reliability and interpretability of predictions. 

Only $14.3\%$ of our surveyed papers (N=9) address the global vs local aspect,  $9.5\%$ (N=6) explore the dynamics, and $1.6\%$ (N=1) tackle framing bias as a comparative task over two or more data samples on the same event. The full list of papers and their categorisation can be found in~\Cref{sec:appendix}.
The remainder of this article links these fundamental disconnects to the more practical research design decisions that arise across both disciplines: data, methods and evaluation.

\section{A Critical Review of Current Practices in NLP and Social Science}
\label{sec:review}
To increase its potential for impact, NLP research needs to reconsider framing and political bias across the entire research pipeline. 
This includes refining benchmarks, methodologies, and evaluation strategies. 
In this section, we make recommendations for each component: exploring new aspects of existing datasets, moving beyond single-label classification and incorporating linguistic features as well as external information, and providing transparent and reliable evaluation outputs with error analyses. We critically compare approaches across NLP and the social sciences, pointing out discrepancies together with practical suggestions for future work.

\subsection{Datasets}\label{sec:data}

\begin{table*}[t]
\small
\centering
\begin{tabular}{@{}llllll@{}}
\toprule
Dataset   & Categories   & Size       & Unit of Analysis         &  Task   \\ \midrule
Bitterlemons   \citep{lin-etal-2006-side}   & Perspective (Israel, Palestine)  & 594        & Documents  &  Classification  \\
Flipper   \citep{chen-etal-2018-learning}     & Left,   Centre, Right                                                                                       &      6,447      & Documents           &  Classification                                                          \\
BASIL   \citep{fan-etal-2019-plain}           & \begin{tabular}[c]{@{}l@{}}Liberal, Conservative,   Centre;\\      Pos, Neu, Neg\end{tabular} &  \begin{tabular}[c]{@{}l@{}}1.2k / 448 \\300       \end{tabular} & \begin{tabular}[c]{@{}l@{}}Spans/Words  \\ Documents        \end{tabular}&  Classification                                                          \\ 
AllSides   \citep{baly-etal-2020-detect}      & Left, Centre, Right                                                                                         & 34k        & Documents           &  Classification                                                          \\
BiasedSents \citep{lim-etal-2020-annotating}	& not-, slightly-, very-, biased & 966	& Sentences  &   Classification            \\
BABE \citep{spinde-etal-2021-neural-media}	& Biased, Non-biased &   3.7k	& Sentences  &   Classification                               \\ 
\begin{tabular}[c]{@{}l@{}}BIGNEWSALIGN\\  \citep{liu-etal-2022-politics} \end{tabular}   & Left, Centre, Right                                                                                         & 1M         & Documents           & Classification                                                          \\
NeuS   \citep{lee-etal-2022-neus}             & Left, Centre, Right                                                                                         & 10.6k      & Documents           &  \begin{tabular}[c]{@{}l@{}}Cross-Doc \\     Summarisation\end{tabular} \\ \midrule 
MFC   \citep{card-etal-2015-media}   & 15 Frames & \begin{tabular}[c]{@{}l@{}}61.5k/ \\  11.9k \end{tabular}  & \begin{tabular}[c]{@{}l@{}}Sentences/ \\      Documents\end{tabular} &   Classification       \\
GVFC   \citep{liu-etal-2019-detecting}        & 9 Frames    & 2.99k      & Headlines  & Classification      \\
\begin{tabular}[c]{@{}l@{}}Multimodal GVFC \\ \citep{tourni-etal-2021-detecting-frames} \end{tabular} & 9 Frames & 1.3k  & \begin{tabular}[c]{@{}l@{}}Headlines \\ $+$ Images   \end{tabular}   & Classification      \\
PVFC \citep{ziems-yang-2021-protect-serve} & \begin{tabular}[c]{@{}l@{}}Entity frames; \\Conservative, Liberal, none     \end{tabular} & 82k  & Documents  & \begin{tabular}[c]{@{}l@{}}Entity frame \\prediction  \end{tabular}\\
\begin{tabular}[c]{@{}l@{}}Narrative Frames \\ \citep{frermann-etal-2023-conflicts}  \end{tabular}  & \begin{tabular}[c]{@{}l@{}} 3 entity roles; 5 frames    \end{tabular} & 428  & Documents  & \begin{tabular}[c]{@{}l@{}}Multi-label \\ frame prediction  \end{tabular}\\
\begin{tabular}[c]{@{}l@{}}SemEval-2023 Task 3 \\ \citep{piskorski-etal-2023-semeval}  \end{tabular}  & \begin{tabular}[c]{@{}l@{}} 14 Generic frames    \end{tabular} & $\sim$1k  & Documents  & \begin{tabular}[c]{@{}l@{}}Multi-label/-class \\ classification   \end{tabular}\\
\bottomrule
\end{tabular}
\caption{Prominent benchmarks for political bias (top) and framing (bottom). We report size (number of data points), unit of analysis, supported task(s) and labels. 
All these data sets are in English and most of them U.S.\ centric.}
\label{tab:datasets}
\end{table*}

Social science studies are characterised by carefully collated data sets which are, however, typically small in size ($\ll$100 articles) and manual labels are rarely released to the public. Hence we focus on limitations and opportunities of NLP framing and bias benchmarks in this section. 
Table~\ref{tab:datasets}, lists relevant datasets, along with details on their labels, size, tasks and unit of analysis. 

\paragraph{Media bias detection} At the \textit{sentence level}, \citet{lim-etal-2020-annotating} used crowdsourcing to annotate sentences on 46 English-language news articles about 4 different events with four levels of bias (not-biased, slightly biased, biased, or very biased). \citet{spinde-etal-2021-neural-media} released BABE (``Bias Annotations By Experts''), a collection of sentences labelled by experts according to binary categories: biased and non-biased, at the sentence and word levels. \citet{fan-etal-2019-plain} provided the BASIL (``Bias Annotation Spans on the Informational Level'') dataset containing sentence (span) and word-level annotations of political leaning and sentiment (stance) towards entities in the article.

At the \textit{document level}, the Bitterlemons corpus \citep{lin-etal-2006-side}, comprises weekly issues about the Palestine--Israel conflict. 
Each issue contains articles from Palestinian and Israeli perspectives written by the portal's editors and guest authors.
Despite being intended for document classification, this dataset can be employed to explore framing and political bias, given the documents' nature of strong bias towards one side of the conflict. 
Additionally, the web portal AllSides\footnote{\url{https://www.allsides.com/about}} categorises news outlets into three political ideologies: right, centre, and left (they also offer a finer-grained five-point scale annotation: left, lean left, centre, lean right, right) with the aim to provide all political perspectives on a given story (cf., Figure~\ref{fig:allsidesimg})
including expert manual assigned categories at the article level. Several research groups have contributed datasets scraped from AllSides~\cite{chen-etal-2018-learning,baly-etal-2020-detect,liu-etal-2022-politics,lee-etal-2022-neus}.

\paragraph{Framing} At the \textit{headline level}, \citet{liu-etal-2019-detecting} released the Gun Violence Frame Corpus (GVFC).
It includes headlines about gun violence in news articles from 2016 and 2018 in the U.S., labelled with frames like politics, economics, and mental health.
\citet{tourni-etal-2021-detecting-frames} released a multi-modal version of the GVFC collection, including the main image associated with each article, and annotations about relevance and framing at the image level. 

At the \textit{document level}, the Media Frames Corpus (MFC, \citealp{card-etal-2015-media}) is the currently most extensive frame-labeled data set available. It includes articles from $13$ U.S.\ newspapers on three policy issues: immigration, same-sex marriage, and smoking. 
This dataset is intended to enable the analysis of policy issue framing, providing annotations at document and span levels with frames like morality, economic, and cultural. More recently, \citet{piskorski-etal-2023-semeval} released a multilingual multifaceted data collection that includes framing as one of the facet with 14 generic framing dimensions at the document level, inspired in the MFC's annotation. 
\citet{ziems-yang-2021-protect-serve} contribute a police violence news articles collection (PVFC) that can be categorised in both domains, media bias and framing. They provide annotations for political leaning: conservative, liberal or none and also entity-centric frames, including the victim's age, race, and gender. 

{\bf Opportunities for Future Work.} In \Cref{sec:casestudy}, we propose three main aspects to investigate framing and media bias. 
(1) \textit{Conducting studies at a local and global level}. \citet{mcleod-etal-2022-navigating} suggest that framing can occur at different textual units in a document. 
Building on this idea, we propose a shift from single label classification on NLP datasets like AllSides, and Bitterlemons. 
As a concrete example, these corpora could be used to identify predictive sentences or spans for particular frames of political biases, and investigate commonalities. This can directly inform social scientists in their analyses as well as tools to expose biases to news consumers.
\citet{roy-goldwasser-2020-weakly} used point-wise mutual information~\citep{church-hanks-1990-word} over bigrams and trigrams to identify spans but found poor generalisation of the approach. \citet{khanehzar-etal-2021-framing} modelled latent frames at the event level, with not explicit validation. Other specific examples with existing data include: exploring the MFC sentence-level annotations to investigate local framing, and then aggregating these labels to gain a global perspective -- an approach that, to our knowledge, has not been done before. Regarding datasets providing sentence-level (BABE) and headline (GVFC) annotation, this can be considered as a local dimension. However, they generalise from the headline to the entire document, which ignores the subtle signals in the local dimension. 
(2) \textit{The dynamics of framing} on various levels are captured by current data sets: the MFC, BASIL, GVFC and BABE provide article timestamps, supporting diachronic modeling of bias and framing. While some studies exist in this domain~\citep{kwak-etal-2020-systematic,card-etal-2022-computational}, the majority of NLP framing considers articles in isolation. Other dynamics, e.g., across countries, communities or media types (e.g., news vs.\ blogs) are of central interest in communication studies but less achievable with existing data sets. Constructing cross-language and/or cross-cultural data sets with articles aligned on the event level is an important first step. 
(3) \textit{Framing as a comparative task}. We propose that researchers explore cross-document differences in their presentation of a specific issue. More concrete, several of the datasets obtained from AllSides include event-level alignment and hence enable comparison across documents on the left--centre--right spectrum at a finer granularity. 

\subsection{Methodologies}
\label{sec:method}

\textbf{In NLP,} researchers have approached media bias as political ideology detection or framing categorisation using different task formulations. 
The first and most common strategy is \textit{single-label classification}, i.e.\ assigning a single label to each data point. At the \textit{word level},  \citet{recasens-etal-2013-linguistic} learn linguistic features from word removal edit-logs in Wikipedia. \citet{spinde-etal-2021-identification} compared the Euclidean distance of word embeddings to identify biased words in articles from Huffington Post (left wing) and Breitbart News (right wing). And \citet{liu-etal-2021-political} experimented with identifying and replacing bias-inducing words with neutral ones using salience scores over word embeddings.

At the \textit{sentence level}, \citet{iyyer-etal-2014-political} used RNNs to identify political ideology in sentences in congressional debate transcripts and articles from the Ideological Book corpus. 
Using the BASIL corpus, \citet{hartmann-etal-2019-issue} correlated sentence and document distributions using a Gaussian mixture model~\citep{reynolds-etal-2009-gaussian} to identify biased sentences;
\citet{chen-etal-2020-detecting} classified biased spans by calculating their probability distributions on news articles;
and \citet{guo-zhu-2022-modeling} applied contrastive learning and created sentence graphs to categorise biased sentences.
Other researchers translated keywords from GVFC into several languages, and fine-tuned mBERT to classify frames in news headlines in languages other than English~\citep{akyurek-etal-2020-multi,aksenov-etal-2021-fine}. 

At the \textit{document level}, there has been substantial work building on the MFC corpus. 
The task has been approached with RNNs~\citep{naderi-hirst-2017-classifying}, attention and discourse information~\citep{ji-smith-2017-neural}, and pre-trained transformer models~\citep{khanehzar-etal-2019-modeling}. 
\citet{baly-etal-2020-detect} combined adversarial adaptation and adapted triple loss with features like Twitter and Wikipedia information about the readers and the outlet to classify the political ideology of news articles. 
More recently, \citet{chen-etal-2020-analyzing} analysed patterns at different granularities (from word to discourse) to identify media bias and \citet{hong-etal-2023-disentangling} developed a multi-head hierarchical attention model to identify biased sentences focusing on their semantic and aggregating those for political bias document classification.
Scholars have performed similar tasks on languages other than English, e.g.\ by translating English keywords in MFC to Russian to investigate the U.S.\ framing in Russian media over 13 years~\citep{field-etal-2018-framing}.

Some work has formalized framing/bias detection as \textit{multi-label classification}, typically adopting unsupervised methods like clustering~\citep{ajjour-etal-2019-modeling} or topic modelling~\citep{tsur-etal-2015-frame,menini-etal-2017-topic} which allows to `softly' assign documents to more than one cluster. 
In a supervised manner, \citet{mendelsohn-etal-2021-modeling} employ RoBERTa to classify multiple framing typologies on immigration-related tweets. Similarly, \citet{akyurek-etal-2020-multi} address multi-label framing over headlines using different configurations of BERT. Both works focus on short documents (headlines or articles capped at 280 characters). The very recent work of \citet{frermann-etal-2023-conflicts} is the first to address document-level multi-label frame classification.
Rather than unstructured, `topic-like' frame detection, some works anchored framing in the depiction of important stakeholders, also referred to as {\it entity framing}~\cite{ziems-yang-2021-protect-serve,khanehzar-etal-2023-probing}.

While we focus on frame and bias {\it detection}, NLP has also proposed methods for \textit{ mitigation}, e.g., by flipping of bias of headlines~\cite{chen-etal-2018-learning} or generating neutral summaries from a collection of biased articles on the same topic~\cite{lee-etal-2022-neus}. These applications come with their own sets of methodological and evaluation challenges, as well as ethical risks, and are beyond the scope of this paper. We advocate for the alternative approach of highlighting frames in multiple articles and presenting them side-by-side as illustrated in \Cref{fig:allsidesimg}, as a safer and potentially more effective approach in raising awareness of bias and framing.

\textbf{In the social sciences,} approaches tend to be manual, with fewer data samples. 
One common approach is to \textit{reason across many documents from a high-level perspective}. 
For example, \citet{chyi-mccombs-2004-media} design and evaluate a two-dimensional framework (spatial and temporal) to investigate framing changes over time in 170 news articles in American English about a U.S.\ school shooting event. 
They manually annotated articles with the signals indicating both of the frame typologies, quantified those annotations and draw conclusions about the temporal and spatial framing behaviour in the inspected articles.
\citet{muschert-2006-media} assessed the previously-proposed framework based on 290 news documents, and confirmed that the present temporal dimension frame still holds when using data from more than one school shooting. 
\citet{hernandez-2018-killed} analysed the framing of 124 news stories from the South China Morning Post (SCMP) about femicides by manually coding the articles and quantifying those observations. The author explored whether those cases were portrayed as isolated cases or part of a systematic social problem, by manually analysing signals like narratives, sources, and the role of the entities.

Communication science studies often \textit{correlate features of news reports with extra-textual information to formulate or validate their hypotheses}.
For example, \citet{mccarthy-etal-2008-assessing} assess media bias in reporting on demonstrations.
They examine media coverage of protests during Belarus's transition from communism, considering factors like protest size, sponsors' status, arrests, and their correlation with media coverage. 
Similarly, \citet{gentzkow-shapiro-2010-what} investigate media bias by calculating think tank citation frequencies in media outlets and correlating them with U.S.\ Congress members mentioning the same groups.

\paragraph{Opportunities for Future Work.} There is a stark disconnect between largely {\it local} approaches to frame modelling in NLP and the  focus on {\it dynamic} and {\it global} questions explored in framing/bias studies in the social sciences. These arguably more complex questions emerging from the social sciences can guide the development of NLP methodologies.
Specifically, capturing subtle signals, including the metaphoric or technical (legal) language use, the correlation with external features, e.g.\ a report's sources, and the broader cultural context in which an article emerged can enrich news framing and bias analysis. Examples at a linguistic level include enriching framing models with notions of metaphoric~\cite{chakrabarty2022s,liu-etal-2022-testing} or subjective~\cite{barron2023clef} language. 
On the cross-document and dynamic level, we propose to address bias and frame classification as a comparative task rather than classifying documents in isolation. This can help {\it induce} frames from data by analysing axes of largest variation; and can naturally support tools and applications to raise readers' bias awareness by exposing them to contrasting perspectives on the same issue. Contextualising framing models with extra-textual, cultural context is arguably the most challenging gap to fill. While it is tempting to suggest the use of large language models to draw some of these connections, we strongly argue for using them at most as an aid for human domain experts, and to scrutinise any automatic predictions due to the known intrinsic biases in these models.

\subsection{Evaluation}
\label{sec:metrics}
We consider two levels of validation: validating data annotations, and validating model predictions.

\paragraph{Validating annotations} Validating the quality of labelled data applies to both the social sciences and NLP. In a typical social science study, the distribution of manual labels is the main factor for accepting or rejecting hypotheses. As such, measures for data quality such as inter-coder reliability (ICR) are routinely reported and a core requisite of the study to ensure that the codebook was correctly conceptualised. Coding often includes discussions and several iterations on trial data~\citep{hernandez-2018-killed}, leading to relatively high ICR scores from carefully trained annotators, often with domain knowledge. For robust NLP model training and validation, reliable annotations are essential. While the assessment of bias or framing are subjective to some extent -- as the assessment of framing depends on the annotator's predispositions -- the development of {\it scalable} annotation frameworks that minimise subjectivity is an important open problem.

\paragraph{Validating (model) predictions} Social science studies are largely analytical examining labelled data, qualitatively based on manual analysis, and quantitatively based on statistical tests. 
In contrast, NLP framing studies primarily rely on empirical methods, evaluating through numerical comparisons with ground truth labels. We propose a shift towards deeper insights, assessing a model's ability to capture framing and political bias on a higher, more abstract level, while also fostering fresh insights into the data. Current approaches fall short of drawing inferences from explicit information, such as assessing story objectivity and factuality. These nuanced, graded strategies require more comprehensive metrics than binary accuracy.

\paragraph{Opportunities for Future Work.} We particularly suggest the consequent adoption of three levels of evaluation: (1) {model performance}, (2) {error analysis}, and (3) {measuring model certainty}. While the three levels are by no means new, NLP work continues to focus on (1), with (2) and (3) given less thought and rigour. NLP research on  media bias would benefit from established standards that guide the error analysis 
as well as measures of model reliability and (un)certainty. Such standards might include reporting of `most challenging' classes and/or instances; categorization of errors; as well as exploring reasons for such short comings~\citep{vilar-etal-2006-error,kummerfeld-klein-2013-error}.
Finally, with the increasing impact of NLP technology on the broader public, users of resulting models (be it news consumers or social science researchers), must have access to model confidence scores to assess the reliability of model predictions, as per point (3).

\section{Discussion}
\label{sec:discussion}
\paragraph{Harmonising depth and scale}
The differences in data sets and evaluation between the disciplines naturally follow from their respective  goals. Framing studies in the social sciences aim to uncover the principles underlying framing and its effects through careful, manual analysis of limited amounts of data, typically grounded in theoretical constructs. The primary goal of NLP in the space of media analysis is automation and scalability. Complex annotation of large training data sets as required for supervised approaches is infeasible. Besides, the required structured annotation paradigms would result in sparse observations of label co-occurrence which in turn would require even larger labelled corpora -- and exploding annotation costs. Harmonising the goal of scalability with depth and theoretical rigour is a difficult problem (that is not specific to the domain of framing and media bias). One approach towards addressing this problem is the use of semi- or unsupervised approaches, which limit the annotations to evaluation sets of more manageable size. Incorporating small amounts of labelled data with powerful pre-trained models is an obvious methodological approach, however, ensuring the validity of predictions and interplay of biases encoded in these models with the target task at hand is an open and important research problem -- particularly in a sensitive domain like media bias analysis.

\paragraph{Feasible yet valid annotation}
How can we obtain ecologically valid annotations in an efficient way and sufficient quantity? We suggest to follow a common strategy in the social sciences: break articles into self-contained segments, on the event or argument level~\citep{muschert-2006-media}.  While recent work on argumentation in online debates has followed a similar approach of segmenting contributions into arguments and annotating frames on the argument level~\cite{ajjour-etal-2019-modeling}, it has not been applied in the news media context. Localised rather than article-level annotations have three advantages: (1)~a cognitively easier task for annotators; (2)~interpretability through the possibility to provide local, extractive evidence for frame predictions; and (3)~a richer document-model of framing that goes beyond the single most likely frame.

\paragraph{Cross-disciplinary expertise for document-external grounding}
Section~\ref{sec:casestudy} pointed to a need for multi-level bias analysis, incorporating local, cross-document and broader cultural contexts. Most NLP work models individual articles without integrating external information or other articles in the collection. A few exceptions exist, including~\citet{baly-etal-2020-detect} who incorporate readership demographics from Twitter and publisher information from Wikipedia; and \citet{kulkarni-etal-2018-multi} who incorporate article link structure into their models. Both works still model data points in isolation, and fall short of incorporating the more subtle cultural, political or societal contexts that inevitably interact with news framing. We argue for a strong role of cross-disciplinarity and human oversight when incorporating those factors, involving domain experts at every step from formulating research questions to model design, transparency, robustness, and evaluation. Cross-disciplinary projects would guide NLP researchers to develop novel methods that are valid and useful for studying the fundamentals of framing and media bias, and equip social scientists with enlarged data sets of high quality and relevance to enrich their research.

\paragraph{Open data} 
NLP has a strong culture of sharing code and annotated data sets to encourage collaboration and reproducibility. This is less common in the humanities. Sharing this data more explicitly through cross-disciplinary dialogue could provide critical assessment and feedback from domain experts. It could drive research into combining large (and potentially noisier) data with small-scale (but high-quality) data sets from the social sciences, to address increasingly complex questions on the emergence and effects of media biases and framing.

\paragraph{The role of NLP in media bias analysis}
Despite a surge in data sets and models for automatic analysis of frames and media bias, the {\it ultimate goal} of these works receives surprisingly little attention. With the broader adoption of NLP methods diverse applications emerge -- from supporting social scientists in scaling their research to larger data samples, to tools that highlight (or even edit) biased news to general public news consumers to expose slanted reporting. An explicit notion of goals and applications (and corresponding statement in research papers) will inform model evaluation, risks and ethical concerns to be discussed in the paper. A mandatory adoption of model cards~\cite{mitchell2019model} is one step in this direction. Irrespective of the final application of NLP research, we argue that NLP can contribute safe and valuable tools and methods only if it recognises the complexity of bias an framing both in its data sets and annotations as well as in its evaluation procedures.

\section{Conclusion}\label{sec:conclusion}
We surveyed recent work in NLP on framing and media bias, and identified disconnects and synergies in datasets, methodologies, and validation techniques to research practices in the social sciences. Despite the opportunities for NLP to support and scale social science scholarship on media bias, a current oversimplification in conceptualisation, modelling, and evaluation limits the validity and reliability of contributions. We have teased out three disconnects and proposed directions for future work, including:
(1) analysing news articles from a local and global perspective, incorporating external non-textual features; 
(2) taking into account the dynamics of framing and bias across documents, cultures or over time; and
(3) tackling the issue of media bias as a comparative task, defining frames on the basis of systematic differences between articles whose origins differ on pre-defined characteristics. This would allow for a more complex characterisation of bias than the currently dominant approach of single-label classification.

\section*{Limitations}
This survey focuses on media bias and `frame building', i.e.\ the manifestation of biases and frames  in news articles. This constrains the scope of our analysis to mainstream print news outlets; and leaves aside the dimension of `frame setting', i.e.\ the effects of those frames on the news consumers. Additionally, we are aware that regardless of the approach taken for sampling the body of previous work included in this paper, given the vast literature in the social sciences, there will be remaining bias in our selection. With the aim of mitigating this bias, we point the reader to complementary surveys in this field, e.g.\ \citet{hamborg-etal-2019-automated} and \citet{ali-hassan-2022-survey}.

\section*{Ethics Statement}
Identifying framing and political bias in news articles is a sensitive application area, and inevitably influenced by social and structural biases in the academic investigators and the pool of annotators.
Datasets and technologies intending to tackle these phenomena comprise the social bias of annotators and researchers developing them in an environment lacking diversity. Besides there is a potential for dual use of models and benchmarks to promote polarisation and misinformation through framing, rather than reduce it. 
We see this paper as an opportunity to identify new directions to diversify NLP methodologies and data sets, grounded in best-practices from the media sciences which have been developed for decades. We anticipate that these steps will, together with a better documentation of models and intended use cases, will help to address the above concerns.

\section*{Acknowledgements}
We thank the anonymous reviewers for their feedback, which significantly improved this paper. We also thank Max Glockner and Vishakh Padmakumar for their constructive suggestions and feedback on this work. 
This article was written with the support from the graduate research scholarship from the Melbourne School of Engineering, University of Melbourne provided to GV. LF is supported by the Australian Research Council Discovery Early Career Research Award (Grant
No. DE230100761).

\bibliography{anthology,GV}

\appendix

\newpage
\section{List of Papers Included}
\label{sec:appendix}
Table~\ref{atable:literature} (on the next page) lists our body of literature, identified as described in Section~\ref{ssec:approach}, and indicates which of our three disconnects are addressed in each paper (if any). The table caption explains our labelling procedure.

\begin{table*}[ht]

\centering
\begin{tabular}{lccc}
\toprule
{\bf Paper}  & {\bf Local/Global} & {\bf Dynamics} & {\bf Comparison} \\ \midrule
\citet{ajjour-etal-2019-modeling}    &              &          &            \\ \hline
\citet{aksenov-etal-2021-fine}       &              &          &            \\ \hline 
\citet{akyurek-etal-2020-multi}      &              &          &            \\ \hline 
\citet{ali-hassan-2022-survey}       &              &          &            \\ \hline 
\citet{baly-etal-2020-detect}       &              &          &            \\ \hline 
\citet{baumer-etal-2015-testing}    &              &          &            \\ \hline 
\citet{cacciatore2016end}            &              &          &            \\ \hline 
\citet{card-etal-2015-media}         &              &          &            \\ \hline 
\citet{card-etal-2022-computational} &              &    x    &            \\ \hline 
\citet{chen-etal-2020-detecting}     & x            &          &            \\ \hline 
\citet{chen-etal-2020-analyzing}     & x            &          &          \\ \hline 
\citet{chen-etal-2018-learning}      &              &          &            \\ \hline 
\citet{chong-druckman-2007-framing}  &              &          &            \\ \hline 
\citet{chyi-mccombs-2004-media}      &              &          &            \\ \hline 
\citet{dallmann-etal-2015-media}    &              &          &            \\ \hline 
\citet{devreese-2005-news}           &              &          &            \\ \hline 
\citet{entman-1993-framing}          &              &          &            \\ \hline 
\citet{entman-2007-framing}          &              &          &            \\ \hline 
\citet{fan-etal-2019-plain}          & x            &         &            \\ \hline 
\citet{field-etal-2018-framing}      &              & x        &            \\ \hline 
\citet{frermann-etal-2023-conflicts} & x            &          &            \\ \hline 
\citet{gentzkow-shapiro-2010-what}   &              &          &            \\ \hline 
\citet{ghanem-1997-filling}          &              &          &            \\ \hline 
\citet{giles-2009-psychology}        &              &          &            \\ \hline 
\citet{gross-2008-framing}           &              &          &            \\ \hline 
\citet{guo-zhu-2022-modeling}        &              &          &            \\ \hline 
\citet{hamborg-2020-media}           &              &          &            \\ \hline 
\citet{hamborg-etal-2019-automated} &              &          &            \\ \hline 
\citet{hartmann-etal-2019-issue}   &              &          &            \\ \hline 
\citet{hernandez-2018-killed}       & x            & x        &            \\ \hline 
\citet{hong-etal-2023-disentangling} & x            &          &            \\ \hline 
\citet{iyyer-etal-2014-political}  &              &          &            \\ \hline 
\citet{ji-smith-2017-neural}       &              &          &            \\ \hline 
\citet{khanehzar-etal-2023-probing} &              &          &            \\ \hline 
\citet{khanehzar-etal-2019-modeling} &              &          &            \\ \hline 
\citet{khanehzar-etal-2021-framing} &              &          &            \\ \hline 
\citet{kiesel-etal-2019-semeval}    &              &          &            \\ \hline 
\citet{kulkarni-etal-2018-multi} &              &          &            \\ \hline 
\citet{kwak-etal-2020-systematic} &              & x        &            \\ 
 
 \bottomrule 
              &                           &              & \emph{Continued on next page}
\end{tabular}
\end{table*}

\begin{table*}[ht]
\centering
\begin{tabular}{lccc}
\toprule
{\bf Paper}  & {\bf Local/Global} & {\bf Dynamics} & {\bf Comparison} \\ \midrule
\citet{lee-etal-2022-neus}  &              &          &            \\ \hline 
\citet{lei-etal-2022-sentence} &              &          &            \\ \hline 
\citet{lim-etal-2020-annotating} &              &          &            \\ \hline 
\citet{lin-etal-2006-side} &              &          &            \\ \hline 
\citet{liu-etal-2021-political} &              &          &            \\ \hline 
\citet{liu-etal-2019-detecting}  &              &          &            \\ \hline 
\citet{liu-etal-2023-things}   &              &          & x          \\ \hline 
\citet{mccarthy-etal-2008-assessing}  & x            &          &            \\ \hline 
\citet{mcleod-etal-2022-navigating} &              &          &         \\ \hline 
\citet{mcquail-2020-mcquails}  &              &          &            \\ \hline 
\citet{mendelsohn-etal-2021-modeling} &              &          &            \\ \hline 
\citet{menini-etal-2017-topic}  &              &          &            \\ \hline 
\citet{muschert-2006-media}  & x            &          &            \\ \hline 
\citet{naderi-hirst-2017-classifying} &             &          &            \\ \hline 
\citet{piskorski-etal-2023-multilingual} &             &          &            \\ \hline 
\citet{recasens-etal-2013-linguistic} &             &          &            \\ \hline 
\citet{roy-goldwasser-2020-weakly} &              &          &            \\ \hline 
\citet{scheufele-1999-framing}     &              &          &            \\ \hline 
\citet{spinde-etal-2021-identification} &           &          &            \\ \hline 
\citet{spinde-etal-2021-neural-media} &             &          &            \\ \hline 
\citet{tourni-etal-2021-detecting-frames} &         &          &          \\\hline 
\citet{tsur-etal-2015-frame}  &              &          &            \\ \hline 
\citet{williams-1975-unbiased} & x            & x        &            \\ \hline 
\citet{ziems-yang-2021-protect-serve}  &              & x        &            \\ \midrule
\textbf{Total}                &      9    &     6   &   1       \\ \bottomrule  
\end{tabular}
\caption{\label{atable:literature}
Cited Literature. Papers marked as `Local/Global' analyse media bias or framing, or provide data at different levels of granularity, ranging from words and sentences (or spans) to entire documents. For a paper to consider `Dynamics', we required the study to include an analysis of the development of a topic across a specific axis, either temporal or spatial (across countries). Papers marked in the `Comparison' column characterise bias or framing by explicitly contrasting data samples from different ideologies or political leanings.}
\end{table*}


\end{document}